\documentclass[conference]{IEEEtran}
\IEEEoverridecommandlockouts
\usepackage[utf8]{inputenc}
\usepackage{cite}
\usepackage{amsmath,amssymb,amsfonts}
\usepackage{algorithmic}
\usepackage[dvipdf]{graphicx}
\usepackage{textcomp}
\def\BibTeX{{\rm B\kern-.05em{\sc i\kern-.025em b}\kern-.08em
    T\kern-.1667em\lower.7ex\hbox{E}\kern-.125emX}}

\begin{document}
\title{\vspace{6mm}Detecting Features of Tools, Objects, and Actions from Effects in a Robot using Deep Learning\vspace{-2mm}
\thanks{$^{1}$N. Saito, K. Kim, and S. Sugano are with Department of Modern Mechanical Engineering, Waseda University, Tokyo, Japan (email: n\_saito@sugano.mech.waseda.ac.jp)}
\thanks{$^{2}$S. Murata is with the National Institute of Informatics, Tokyo, Japan and SOKENDAI (the Graduate University for Advanced Studies), Tokyo, Japan}
\thanks{$^{3}$T. Ogata is with Department of Intermedia Art and Science, Waseda University, Tokyo, Japan, and National Institute of Advanced Industrial Science and Technology (AIST), Tokyo, Japan}}
\author{\IEEEauthorblockN{Namiko Saito$^{1}$, Kitae Kim$^{1}$, Shingo Murata$^{2}$, Tetsuya Ogata$^{3}$, and Shigeki Sugano$^{1}$}
}
\maketitle

\begin{abstract}
We propose a tool-use model that can detect the features of tools, target objects, and actions from the provided effects of object manipulation. We construct a model that enables robots to manipulate objects with tools, using infant learning as a concept. To realize this, we train sensory-motor data recorded during a tool-use task performed by a robot with deep learning. Experiments include four factors: (1) tools, (2) objects, (3) actions, and (4) effects, which the model considers simultaneously. For evaluation, the robot generates predicted images and motions given information of the effects of using unknown tools and objects. We confirm that the robot is capable of detecting features of tools, objects, and actions by learning the effects and executing the task. 
\end{abstract}

\section{Introduction}
In recent years, robots have become part of human living space and have been expected to perform various tasks in complex environments. If robots could use tools as humans do, they could improve in versatility, overcome some physical limitations and adapt to the environment. Therefore, research on tool-use by robots has aimed at robots that are useful for daily life.\\
Piaget \cite{Piaget} proposed that infants learn tool-use as part of the cognitive developmental process as detailed in \cite{b1}, \cite{b2}. In the first to third substages of Piaget's sensorimotor developmental stage (0--8 months), infants come to understand motor behavior and learn to couple perception and action. Next, in the fourth substage (8--12 months), they learn the causal relationship between actions and effects. In other words, they become able to understand what results from their actions. Then, in the fifth substage (12--18 months), they acquire the features of tools and objects by trial-and-error of object manipulation with tools. The features are unique to each combination and can be inferred from the causality between the actions and the effects. Finally, in the sixth (and final) substage (18--24 months), they become able to make plans for tool-use, even without conducting actions. They can understand how they should act and what they should use, understanding this from the effects. In other words, they become able to detect proper features of tools, objects, and actions from observed effects.\\
Despite this development process being known, most conventional research on robot tool-use has focused on either tools or objects. Nishide et al.~\cite{b3} and Montesano et al.~\cite{b4} enabled robots to learn the relations between actions and object movements. Along another line, Stoytchev~\cite{b5}, Nishide et al.~\cite{Nishide}, Mar et al.~\cite{Mar}, and Takahashi et al.~\cite{b6} enabled robots to acquire tool function for a specific target object. In each of these studies, robots did not consider tools, objects, and actions simultaneously, the way humans do. A study by Goncalves et al.\cite{b7} focused on both tools and objects, but they aimed to determine features of tools and objects on the basis of categories, such as area and circularity, set in advance by the experimenters. Therefore, it was difficult for the robot to acquire the features autonomously and manipulate arbitrary objects with arbitrary tools without requiring human assistance. \\
In this study, we refer to Piaget's developmental process and use this analogy to enable a robot to detect the correct tools, objects, and actions from information on effects. To realize this, it is necessary to consider four factors: (1) tools, (2) target objects, (3) actions, and (4) effects. These must be considered simultaneously because the effects will change if one of the other three factors changes. For example, a pulling action targeting a ball will result in motion if the tool is a rake but not if the tool is a stick. Therefore, we set tasks that include all four factors and have the robot experience them. Then, we construct a model that can consider the four factors simultaneously. \\
This paper is organized as follows. Section I\hspace{-.1em}I describes the method to construct the tool-use model. Section I\hspace{-.1em}I\hspace{-.1em}I presents the experimental setup and explains how the four factors are considered simultaneously. Section I\hspace{-.1em}V presents the experimental results, and Section V discusses them. Finally, Section V\hspace{-.1em}I concludes.

\section{Tool-use Model}
\begin{figure}[tbp]
\centerline{\includegraphics[width=7.7cm]{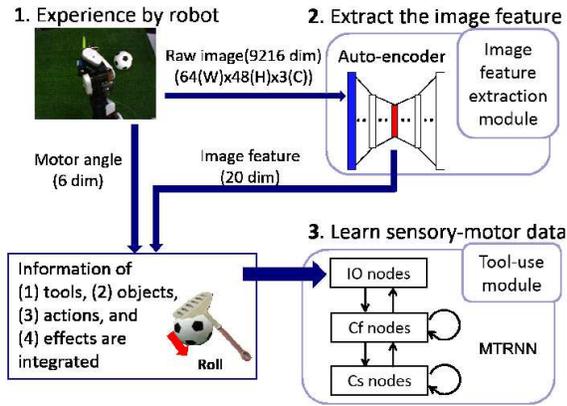}}
\caption{Proposed model, which can consider all four factors at the same time. This model is constructed and used in three steps. First, a robot experiences some tasks and records sensory-motor data during the experience. Second, image features are extracted by the image-feature extraction module. Third, the image features and motor information are integrated and used to train the tool-use module. MTRNN: multiple time-scale recurrent neural network}
\label{model}
\end{figure}

In this section, we describe the tool-use model, which can consider the four factors simultaneously. Figure \ref{model} shows an overview of the model based on the study of Takahashi et al. \cite{b6}. The model contains two modules: an image feature extraction module and a tool-use module. The model is constructed and used as follows.
\begin{enumerate}
 \item A robot experiences some object manipulation with tools, deriving motion and visual information from this. 
 \item The visual information is trained by applying the image-feature extraction module, and the visually useful elements are extracted as image features.
 \item Time-series data of the image features and motion information are integrated. Then, the integrated data are fed to the tool-use module and learned as integrated information.
\end{enumerate}

\subsection{Tool-use Experience by a Robot}
A robot experiences several tasks in combination of multiple tools, objects and actions. The joint angles of the robot arm are recorded as motion information and images taken by a camera mounted on the robot are recorded as visual information. It is important that the tasks display differences between effects, such as object falling, sliding, and failing to move. Choosing tasks like this allows the recorded sensory-motor data to include information about all four factors.
\subsection{Extraction of Image Features}
For the image-feature extraction module, we use Convolutional Auto-Encoder (CAE) \cite{CAE}. The Auto-Encoder (AE) \cite{Hinton} technique uses a structure that narrows in the middle, like an hourglass. The input data passes through this middle layer with a small number of nodes, and then the processed data are output with the original dimensionality. Since the output data are learned with the aim of reconstructing the input data, it is possible to extract features of images from only a few dimensions in the middle-layer nodes. CAE is AE with a convolution layer. The raw images taken by the robot are compressed by CAE, and the features of tools, objects, and robot-arm movement shown in the images are processed in low dimensions. Because of the good generalization offered by CAE, it is possible to represent unknown tools and objects as image features.
\subsection{Learning Sensory-motor data}
For the tool-use module, we use a multiple timescale recurrent neural network (MTRNN) \cite{MTRNN}, a type of recurrent neural network that can predict the next state from a current state. MTRNN uses three types of node, and these types different by time constant: input--output (IO) nodes, fast context (Cf) nodes, and slow context (Cs) nodes. The Cf nodes learn primitives of movement in the data, whereas the Cs nodes learn sequences in the data. By combining the three node types, the dynamics of time-series data can be learned. This module learns to integrate the time series of image features and motor angles. \\
In forward calculation, the output value is computed. First, the internal value of the \(i\)th neuron \(u_i\) at step \(t\) is calculated as
\begin{equation}
u_i(t)=\biggl(1-\frac{1}{\tau_i}\biggr)u_i(t-1)+\frac{1}{\tau_i}\left[\sum_{j\in N}w_{ij}x_j(t)\right],
\end{equation}
where \(N\) is the index sets of neural units, \(\tau_i\) is the time constant of the \(i\)th neuron, \(w_{ij}\) is the weight of the connection between the \(j\)th and \(i\)th neuron, and \(x_j(t)\) is the value input to the \(i\)th neuron by the \(j\)th neuron. Then, the output value is calculated by 
\begin{equation}
y_i(t) = tanh\left(u_i(t)\right). 
\end{equation}
The value of \(y_i(t)\) is used as the next input value: 
\begin{equation}
x_i(t+1)=y_i(t).
\end{equation}
In backward calculation, we use the back-propagation through time (BPTT) algorithm \cite{BPTT} to minimize the training error in \eqref{y}, and update the weights by \eqref{weight}. 
\begin{equation}
E=\sum_{i}\sum_{i\in{IO}}\left(y_i(t-1)-T_i(t)\right)^2\label{y}
\end{equation}
\begin{equation}
w^{n+1}_{ij}=w^{n}_{ij}-\alpha \frac{\partial E}{\partial w^{n}_{ij}}
\label{weight}
\end{equation}
Here, \(T_i(t)\) is the \(i\)th input given as teaching data, \(\alpha\) is the learning rate, and \(n\) is the number of iterations. At the same time, the initial value of the Cs layer (Cs(0)) is also updated to store the features of the dynamics information, as
\begin{equation}
Cs^{n+1}_i(0)=Cs^n_i(0)-\alpha \frac{\partial E}{\partial Cs^n_i(0)}.\label{Cs}
\end{equation}
Thanks to this, it is expected that each feature of the four factors will be self-organized in Cs(0) space. In other words, the features of the four factors are accumulated in Cs(0) space. \\
In addition, when dealing with unknown tools and objects, the Cs(0) value that best matches the task can be calculated from a trained network. This process is called recognition. The dynamics information detected by the robot is embedded in the recognized member of Cs(0). Therefore, by inputting the value of Cs(0) to the trained network with initial input data (IO(0)), it is possible to generate a sequence of predicted images for the behavior and motions corresponding to the value of Cs(0). We can evaluate whether the model has the ability to detect the features by examining the generated result.
\section{Experimental Setup}
\subsection{Task Design}
\begin{figure}[tbp]
\centerline{\includegraphics[width=8cm]{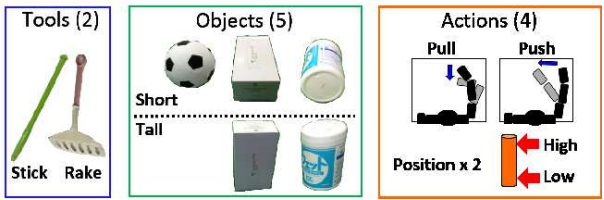}}
\caption{Tools, objects, and actions for training. Tools are rake- or stick-shaped. Objects are a ball, a tall box, a short box, a tall cylinder, and a short cylinder. Actions are pull (high and low) and push (high and low).}
\label{setup}
\end{figure}

To guide a robot to experience information that includes all four factors, it is necessary to use experiments that have a variety of combinations of tool, object, and action. We used a humanoid robot, NEXTAGE developed by Kawada Robotics, and guided it to try some tasks with its right arm, which has 6 degrees of freedom. As shown in Fig. \ref{setup}, we prepared two kinds of tool (a stick shape and a rake shape) and three kinds of target object (a ball, a tall box, a short box, a tall cylinder, and a short cylinder). For actions, we used four kinds: pushing sideways and pulling toward the robot, with each action performed at either a high or low position. The robot was not instructed to pull tall objects at a low position because doing so might cause destruction of the tools or objects and there was a risk of imposing a high load on the robot. Therefore, we set a total of 36 tasks (=\(2\times5\times4-4\); tools \(\times\) objects \(\times\) actions $-$ forbidden cases). In all tasks, the robot began with the tool gripped, and we set the initial joint angles and objects in the same way each time. 
\subsection{Setting for Training}
\begin{figure*}[tbp]
\centerline{\includegraphics[width=12.cm]{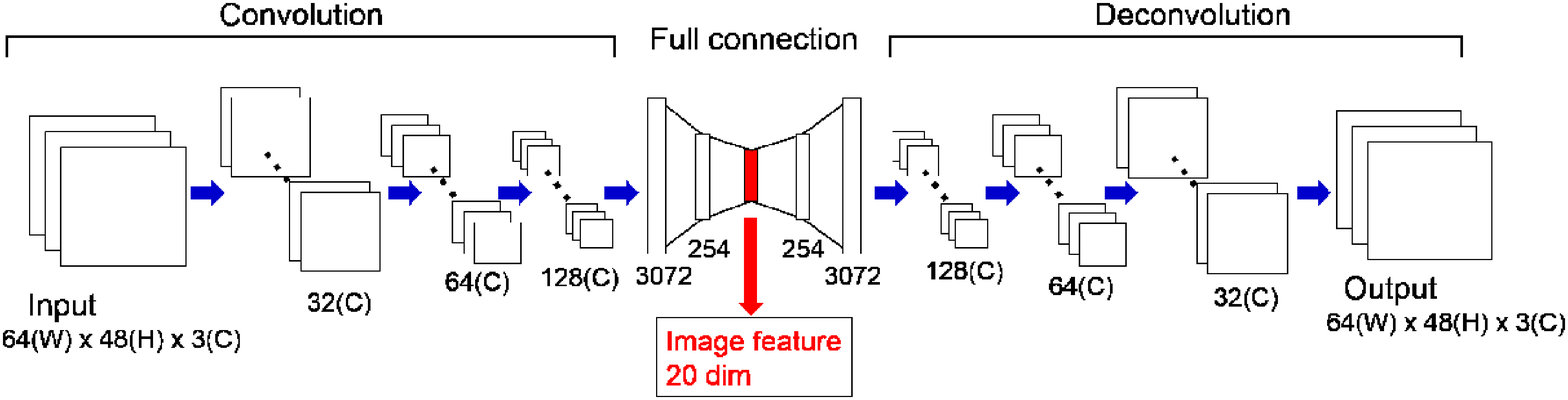}}
\caption{Structure of CAE. CAE consists of convolution layers and full connection layers. This module compressed 9216-dimensional raw images to 20-dimensional image features.}
\label{CAE}
\end{figure*}

The time length to perform the 36 tasks ranged from 7.5 to 14.4 s. By keeping the robot stationary at the final position after task completion, we recorded sensory-motor data for 14.4 s in each task, sampling each 0.1 s (144 frames for each action). Among the sensory-motor data, the visual information had 9216 dimensions (64 \(\times\) 48 \(\times\) 3; width \(\times\) height \(\times\) channel). The high-dimensional data were compressed to 20 dimensions by CAE. Figure \ref{CAE} shows the structure of CAE. The extracted 20-dimensional image features and 6-dimensional joint angle data were put into MTRNN. The values of the extracted image features were rescaled to [-1, 1], and joint angle data were rescaled to [-0.8, 0.8]. Table \ref{tab1} shows the structure of MTRNN. 
\begin{table}[t]
\caption{Structure of MTRNN}
\begin{center}
\begin{tabular}{|c|c|c|}
\hline
\textbf{Node Name} & \textbf{ Number of Nodes}& \textbf{Time Constant} \\
\hline
\textbf{IO nodes} & 26& 1 \\
\hline
\textbf{Cf nodes} & 50& 5 \\
\hline
\textbf{Cs nodes} & 6& 40 \\
\hline
\end{tabular}
\label{tab1}
\end{center}
\end{table}
\subsection{Experimental Evaluation}
\begin{figure}[tbp]
\centerline{\includegraphics[width=5cm]{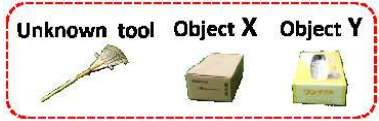}}
\caption{Untrained rake-shaped tool and box-shaped objects. Evaluation experiments A and B were conducted using them.}
\label{unknown}
\end{figure}

For model evaluation, we evaluated whether the model can well predict images and motions of tool-use tasks involving ``unknown'' tools and objects. If so, the trained model is regarded as detecting the features of tools, objects, and actions.\\
For the recognition task in Cs(0), we provided the initial and final states of unknown tasks to the model. At this time, the final state included only the image data. These data show the effects because behavior of the object can be understood by comparing the initial and goal images. Cs(0) was calculated as \eqref{Cs} such that the error between the target image and the image at the final step obtained by repeating next-state prediction from the initial state was progressively smaller. At this time, the error was calculated using the trained MTRNN as follows.
\begin{equation}
E=\left\{ \begin{array}{lll}
\sum_{i\in{IO}}\left(y_i(t-1)-T_i(t)\right)^2 & t=1\\
\sum_{i\in{img}}\left(y_i(t-1)-T_i(t)\right)^2 & t=144\\
0 & otherwise\\ 
\end{array}\right.
\end{equation}
 Evaluation experiments were conducted using two unknown short boxes and an unknown rake, shown in Fig. \ref{unknown}. We prepared two experiments, A and B, and performed the experiments using the objects. 
\begin{description}
\vspace{1pt}
 \item[Experiment A]\mbox{}\\
    \hspace{-10pt}The robot performs ``pull low'' with the tool, and the object ``slides'' to the front of the robot.
 \item[Experiment B]\mbox{}\\
    \hspace{-10pt}The robot performs ``pull low'' with the tool, but the object ``does not move.''
\vspace{1pt}
\end{description}
We put initial state and goal images of each experiment A and B to the model. Since the actual task execution of experiment B is impossible, the goal images of this experiment were artificially prepared. We set the same tool and object and showed the same arm position in the images, but the robot was expected to detect different features due to the difference of the effects. In experiment A, we expected the robot to detect the tool as ``rake,'' which can make marks, and the object as ``short object,'' which can slide without falling over. In experiment B, there were two patterns of proper detection. First, the robot could detect the action as ``low pull,'' the tool as ``stick'' (which cannot make marks), and the object as ``short object,'' which does not fall over even if it is hit by a tool. Second, the robot could detect ``high pull'' paired with ``short object'' because the object would not be moved by this combination of action and object. In this pattern, the detected tool is expected to be more likely to be ``rake'' because it has the least difference from the setting. In Table \ref{expect}, we summarize all four factors' relations suggested above. The detection of tool, object, and action is said to be correct if the detected combination satisfies this condition. By putting the recognized Cs(0) and initial state (IO(0)) into the trained model, the robot predicted images and motions. We evaluated the robot's detection from the generated results. We also checked whether the generated images and motions connect the first and the final state sequentially, considering the movement of robot arm (i.e., the action), the tool, and the object. 
\begin{table}[t]
\caption{Expected Combinations of Four Factors}
\begin{center}
\begin{tabular}{|c|c|c|c|c|}
\hline
\textbf{Experiment} & \textbf{Tool}& \textbf{Object}& \textbf{Action}& \textbf{Effect} \\
\hline
A &Rake&Short object&Low pull&Slide \\
\hline
B & Stick& Short object&Low pull&No movement \\
\hline
B & Rake&Short object &High pull& No movement\\
\hline
\end{tabular}
\label{expect}
\end{center}
\end{table}
\section{Experimental Results}
\begin{figure}[tbp]
\centerline{\includegraphics[width=7.5cm]{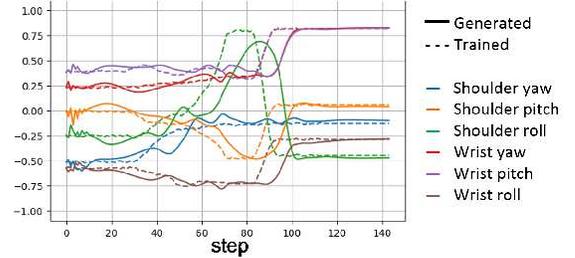}}
\caption{Trajectory of the robot arm six joint angles in experiment A using tool X. The data are rescaled to [-1, 1]. The dotted lines show the trained angle data for the ``low pull'' action, which is a target for generation. The solid lines show the motor angle data generated by using the recognized Cs(0).}
\label{angle}
\end{figure}
\begin{figure*}[tbp]
\centerline{\includegraphics[width=15.5cm]{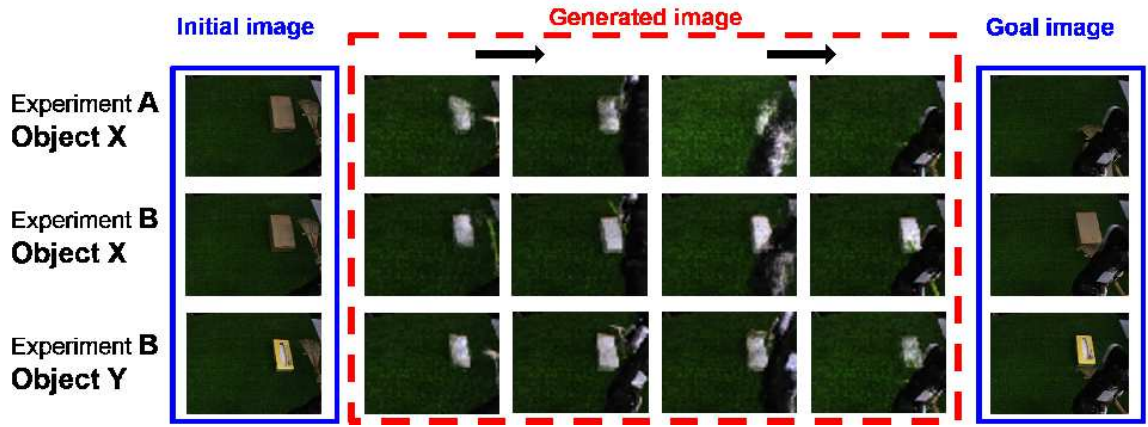}}
\caption{Figures in blue squares show the given initial images and goal images. Figures in red square show the predicted image data generated by the robot. In order from the top row, we show the result of experiment A with object X, experiment B with object X, and experiment B with object Y. In experiment A, the generated images show the sequential task of drawing the object, and in experiment B, the generated images show the tasks that do not move the objects. In experiment A and experiment B with object Y, the generated figures show the tool as ``rake'', whereas in experiment B with object X, the generated figures show the tool as ``stick.''}
\label{imggen}
\end{figure*}
\begin{figure*}[tbp]
\centerline{\includegraphics[width=15cm]{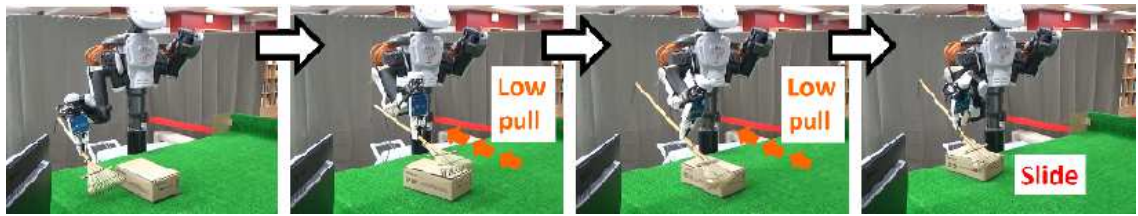}}
\caption{Transition of the robot's movement of experiment A with object X. The joint angles are generated with trained MTRNN using recognized Cs(0). The robot pulled the box with the rake in the low position and succeeded in drawing it in front.}
\label{gen}
\end{figure*}
\begin{figure*}[tbp]
\centerline{\includegraphics[width=15cm]{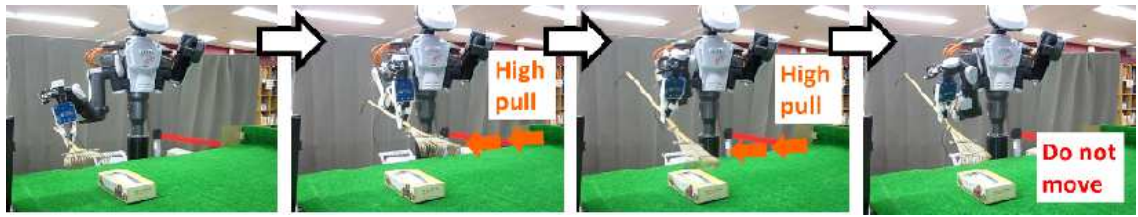}}
\caption{Transition of the robot's movement of experiment B with object Y. The robot conducted ``high pull'' action with the rake. Therefore, the box did not move. The result of the object behavior matched the goal image. }
\label{gen2}
\end{figure*}
\begin{figure}[tbp]
\centerline{\includegraphics[width=7cm]{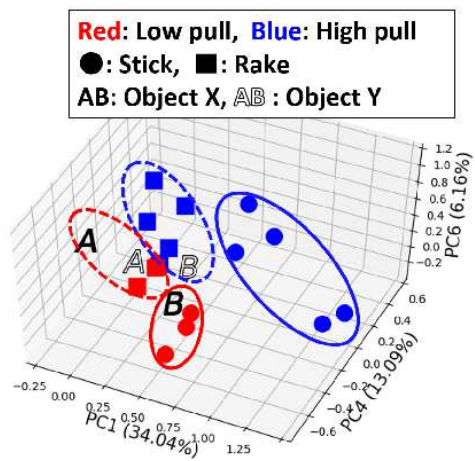}}
\caption{PCA of Cs(0) of trained tasks. PC1, PC4, PC6 are shown. We plot 16 ``pull'' tasks with a rake and a stick. Different features of pulling height and tools are separated and clustered. In addition, we plot PCA of Cs(0) of experiment A and B on the space.}
\label{actool}
\end{figure}

We trained MTRNN with the data of 36 tasks, performing this 150,000 times. Then, we performed Cs(0) recognition 150,000 times for each of the two experiments using the initial and final states as in the blue squares in Fig. \ref{imggen}. Using the recognized Cs(0) value, the robot generated predicted images and motions. Fig. \ref{angle} shows, as an example of the training data and generated result, the trajectory of the robot arm's motor angles using object X in experiment A. The dotted lines show the training data and the solid lines show the data generated by using the recognized Cs(0). Although the $x$-axis (i.e., time) direction of the solid lines is shifted, the $y$-axis direction (i.e., angle) is close to the target dotted lines. In other words, it was possible to generate data to recreate the target operation, though with different timing.\\
The top row of Fig. \ref{imggen} shows the predicted images in experiment A using unknown object X. We can see that the robot properly drew a short box pulled forward with a rake. Therefore, we can say the robot detected the combination of the four factors as ``rake,'' ``short box,'' ``low pull,'' and ``slide,'' which matches the requirements shown at the top row of Table \ref{expect}. We also confirmed that the robot generated the motor angles to actually move according to the data. As a result, the robot was able to perform the task properly and succeeded in sliding the box with the rake, as shown in Fig. \ref{gen}. The result using object Y was similar.\\ 
The predicted images for experiment B using unknown object X are shown in the middle row of Fig. \ref{imggen}. As can be seen, the robot predicted a ``low pull'' action with a ``stick,'' and the tool passed over the ``short box.'' This combination matched the requirements shown for the case in Table \ref{expect}. Actual task execution by the robot is not shown because this experimental setting cannot be tested in the real environment. Similarly, the predicted images for experiment B using unknown object Y are shown in the bottom row of Fig. \ref{imggen}. These images show that the robot conducted a ``pull'' action with a rake, and the tool passed over the ``short box.'' The robot also generated the motor angles and to actually move according to the data, as shown in Fig. \ref{gen2}. As can be seen, the object did not move because the robot pulled at the high position. Therefore, the robot detected the combination of the four factors as ``rake,'' ``short box,'' ``high pull,'' and ``no movement'', which matched the requirements shown in the bottom row of Table \ref{expect}. \\
In addition, we carried out principal component analysis (PCA) of the Cs(0) value of the trained tasks to confirm whether the robot could self-organize features. We select three axes that are easy to see and plotted the result of 16 ``pull'' tasks in Fig. \ref{actool}. We indicate different position of the action by different colors and different tools by different markers. As can been seen, clustering and separation distinguish among different features. In other words, the robot acquired the features of actions and tools properly. We also analyzed the detection described above by PCA of Cs(0) from experiments A and B, then plotted the results on the trained tasks' Cs(0) space. As shown in Fig. \ref{actool}, each experiment's Cs(0) is located as follows. Two plots of A are located in the cluster of ``low pull'' and ``rake.'' The plot of B using object X is located in the cluster of ``low pull'' and ``stick.'' The plot of B using object Y is located in the cluster of ``high pull'' and ``rake.'' In all experiments, every plot is located exactly in the cluster that presents the detected feature's combination of actions and tools as mentioned above. This analysis confirms reliable recognition.
\section{Discussion}
In contrast with previous studies, we constructed a model that can simultaneously consider four factors of object manipulation (tool, target object, action, and effect) and thereby acquire combined features  according to the trained experience. Human infants develop this process during the fifth substage of Piaget's sensorimotor developmental stage. As shown in Section I\hspace{-.1em}V, the detected combinations of the four factors matched the expectations shown in Table \ref{expect} for the experiments. Note that in experiments A and B with object X, the same tool was used, but the robot recognized different tool features from the effect of the operation. In addition, comparing  experiments A and B using object Y, the robot conducted different actions in response to the different effects despite the robot arm having the same placement in the input images. This indicates that the robot could detect the proper features of tools, objects, and actions from the effects, even without concretely performing the task. In addition, we showed that the robot could generate sequences of predicted images. This is similar to the effect of infants mentally planning actions. We also showed that the model could generating motions, which enabled the robot to execute tasks correctly. In other words, the robot could reproduce the effect from the detected features. This is one of the abilities acquired by infants during the sixth substage of Piaget's sensorimotor developmental stage. 
\section{Conclusion}
In this study, we proposed a tool-use model that can detect tools, objects, and actions. This model mimics infant developmental of similar ability. We set tasks that include information about four factors (tools, objects, actions, and effects), and the robot experienced them. We trained the sensory-motor data recorded during the experience with a tool-use model consisting of an image-feature extraction module and a tool-use module. The results showed that it was possible to detect the appropriate features from an initial and final state of a task using unknown tools and objects. The robot succeeded in executing actual tasks with generated joint angles and properly generated predictive images. In future work, we plan to construct a model that can select proper tools and conduct proper actions according to the state at the moment of choice and the final state. This model will enable a robot to behave according to the position of the object and its arm, which would improve versatility and ability to cope with more complicated tasks.

\section*{Acknowledgments}
This work was supported in part by a JSPS Grant-in-Aid for Scientific Research (S) (No. 25220005), a JST CREST Grant (No. JPMJCR15E3), and the ``Fundamental Study for Intelligent Machine to Coexist with Nature'' program of the Research Institute for Science and Engineering, Waseda University, Japan.

\end{document}